\documentclass[runningheads]{llncs}
\usepackage{graphicx}
\usepackage{bbding}
\usepackage{amsmath}
\usepackage{cite}
\usepackage{multirow}
\usepackage{graphics} 
\usepackage{epsfig} 
\usepackage{mathptmx} 
\usepackage{times} 
\usepackage{amsmath} 
\usepackage{amssymb}  
\usepackage{cite}
\usepackage{multirow}
\usepackage{booktabs}
\usepackage[T1]{fontenc} 
\usepackage{enumerate}

\begin{document}
\title{Inferior Alveolar Nerve Segmentation in CBCT images using Connectivity-Based Selective Re-training}
%
%
\author{Yusheng Liu \and
Rui Xin\and
Tao Yang\and
Lisheng Wang\inst{(}\Envelope\inst{)}}
\authorrunning{F. Author et al.}
%
\institute{Institute of Image Processing and Pattern Recognition, Department of Automation,
Shanghai Jiao Tong University, Shanghai 200240, People’s Republic of China\\
\email{\{lys\_sjtu,xr1999,yangtao22,lswang\}@sjtu.edu.cn}}
\maketitle              
\begin{abstract}
    Inferior Alveolar Nerve (IAN) canal detection in CBCT is an important step in many dental and maxillofacial surgery applications to prevent irreversible damage to the nerve during the procedure.
    The ToothFairy2023 Challenge aims to establish a 3D maxillofacial dataset consisting of all sparse labels and partial dense labels, and improve the ability of automatic IAN segmentation.
    In this work, in order to avoid the negative impact brought by sparse labeling, we transform the mixed supervised problem into a semi-supervised problem. Inspired by self-training via pseudo labeling, we propose a selective re-training framework based on IAN connectivity.
    Our method is quantitatively evaluated on the ToothFairy verification cases, achieving the dice similarity coefficient (DSC) of 0.7956, and 95\% hausdorff distance (HD95) of 4.4905, and wining the champion in the competition. 
    Code is available at https://github.com/GaryNico517/SSL-IAN-Retraining.

\keywords{Self-training\and Inferior alveolar nerve segmentation\and Semi-supervised learning.}
\end{abstract}
\section{Introduction}
Cone beam computed tomography (CBCT) is becoming increasingly important for treatment planning and diagnostics in dental applications such as orthodontics, dental implants, and tooth extractions \cite{abesi2023accuracy}. 
Three-dimensional parsing of the inferior alveolar nerve (IAN) from CBCT is crucial \cite{Cipriano_2022_CVPR,cipriano2022deep} in extraction of impacted teeth or removal of cystic lesions, preventing important anatomical or neural structures from irreversible damage.

The ToothFairy Challenge \cite{Cipriano_2022_CVPR} aims to advance the development of deep learning frameworks to segment the IAN from publicly available 3D sparsely and densely annotated CBCT datasets.
In recent years, a large amount of research has been devoted to the development of automatic IAN segmentation in CBCT scans. 
Traditional methods use statistical shape models (SSM) \cite{abdolali2017automatic,kainmueller2009automatic} or threshold segmentation methods \cite{moris2012automated}, but these techniques rely on additional manual work by experts, and differences in image distribution can also result in incomplete segmentation of neural structures. 
The existing deep learning method \cite{jaskari2020deep} is designed based on CNN, but the rough annotation conceals the true performance of the method.

In order to avoid the time-consuming and labor-intensive fine-grained voxel labeling, semi-supervised semantic segmentation is proposed to learn the model from a small number of labeled images and a large number of unlabeled images. 
Self-training \cite{yang2022st++,zhang2022self} is often considered a form of entropy minimization in semi-supervised learning (SSL), since the re-trained students are supervised with pseudo-labels trained by the teacher on labeled data. 
However, when using these inappropriate pseudo-labels to iteratively optimize the model, it may lead to performance degradation. 
As a result, the pseudo-label selection strategy directly determines the performance of the model.

In this paper, we transform the challenging mixed-supervised task into a semi-supervised task without using sparse annotations in the dataset. 
The contributions of our work are summarized as follows:
\begin{enumerate}
    \item We adopt an advanced nnUNet-based self-training framework to perform selective iterative training by screening and prioritizing according to the connectivity characteristics of IAN and the stability of prediction results during model training.
    \item We design strong data augmentation (SDA) \cite{zhang2022self} and test-time augmentation (TTA) on non-densely labeled images before iterative training to mitigate overfitting noisy labels and decouple similarity between teacher and student predict.
    \item Our method won the first place in Miccai ToothFairy Challenge, showing good segmentation performance.
\end{enumerate}
\begin{figure*}[t]
	\centering
	\includegraphics[width=5in]{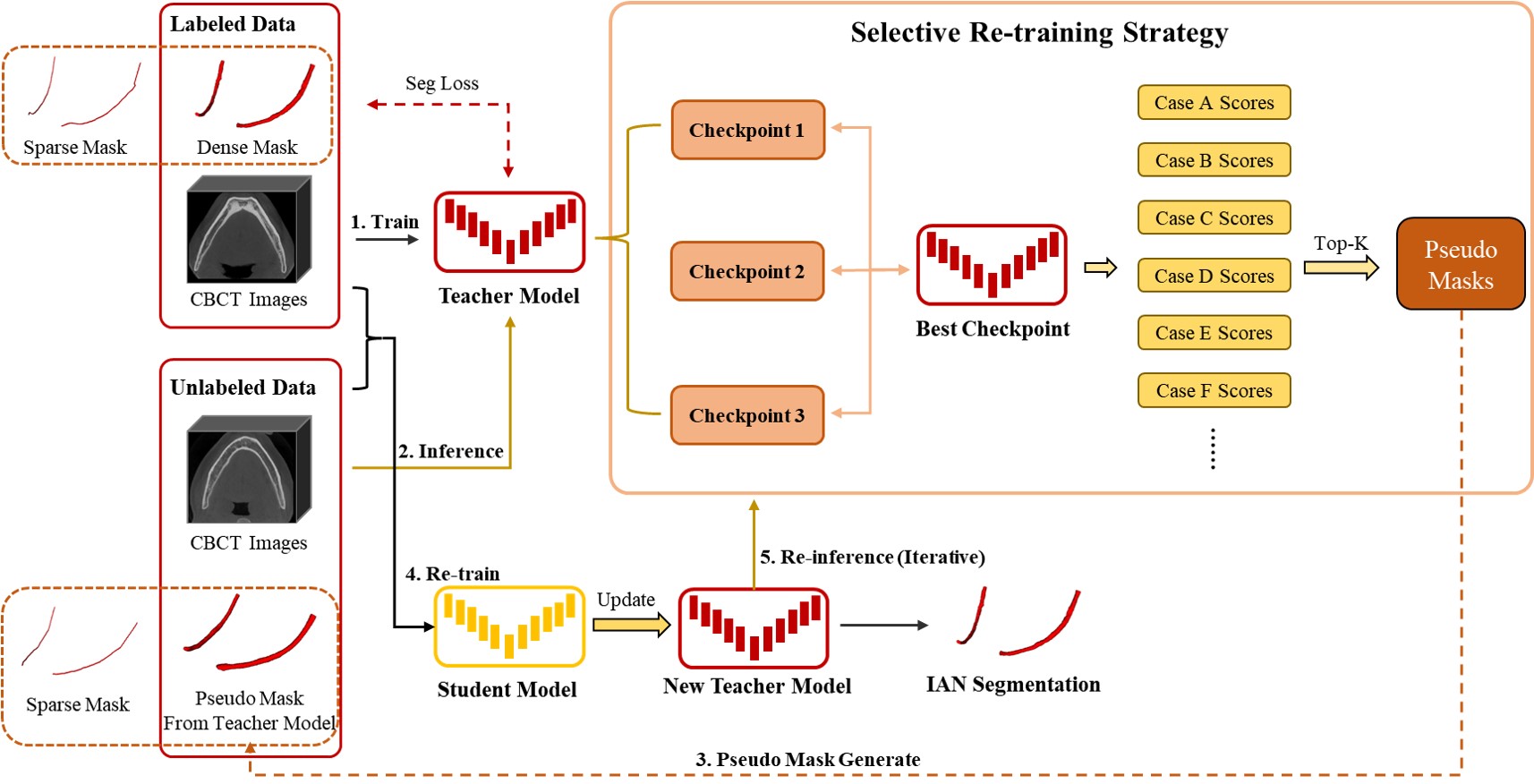}
	\caption{The overview of our framework. The framework does not use sparse annotations in the dataset and consists of teacher models and student models. 
    The teacher model trained by limited densely labeled data generates pseudo-labels. 
    After the pseudo-labels are screened out through the selective re-training strategy, the student model is trained together with the labeled data and data with pseudo masks, and then updated to the teacher model in the next iteration.}
	\label{framework}
\end{figure*}

\section{Methods}
The pipeline of the proposed segmentation framework is shown in Fig.\ref{framework}. 
We adopt nnUNet \cite{isensee2021nnu} as the basic network models for IAN segmentation. 
Self-training method is adopted for semi-supervised semantic segmentation. 
In addition, a connectivity-based selective re-training strategy is designed to screen more plausible pseudo-labels. 
A detailed description of the method follows.

\subsection{Prepropcessing}
The proposed framework includes the following preprocessing steps:
\begin{itemize}
    \item[$\bullet$] Labeled data filtering: In ToothFairy dataset, some dense labels are incomplete while a normal mandibular canal should not. 
    This happens due to a lack of density difference in such areas of the jawbone, this can be due to CBCT acquisition noise or patient specific conditions. 
    
    In order to avoid misleading model learning by disconnected labeling information, we perform connected component labelling (CCL) on the dense labels in the dataset, and filter out labels with more than 2 connected domains.
    
    In addition, for the case where the boundaries of IAN at both ends are blurred, we calculate the difference between the model prediction results and the Dense annotations in the training set, filter out the data with DSC lower than 0.85, and only select the data with definite boundaries.
    \item[$\bullet$] Resampling: All the images are resampled to the same target spacing [1, 1, 1] by using the linear spline interpolation so that the resolution of the z-axis is similar to that of the x-/y-axis. 
    The patch size of the teacher and student model input are [80, 160, 192].
    \item[$\bullet$] Intensity normalization: A z-score normalization is applied based on the mean and standard deviation of the intensity values from the foreground of the whole dataset
\end{itemize}

\subsection{Proposed framework}
Inspired by self-training framework (namely ST++ \cite{yang2022st++}),  our framework generates plausible pseudo-labels in semi-supervised semantic segmentation task. We employ nnUNet for both teacher and student models. 
The detail of our framework is as follows:
\subsubsection{Selective re-training strategy.}

We adopt a selective re-training scheme to expand the labeled data in samples by prioritizing the reliability of unlabeled samples. 
A measure of reliability or uncertainty on unlabeled images is to compute the overall stability of the evolving pseudo-mask across iterations throughout training. 
Therefore, more reliable and better predictive unlabeled images can be selected according to their evolutionary stability during training.

As shown in Fig.\ref{framework}, several model checkpoints are saved while training the teacher model with densely labeled data supervision, and the difference in their predictions for unlabeled images serves as a measure of reliability. 
As the trained model tends to converge and reach the best performance late in training, we evaluate the average Dice between each early pseudo mask and the final mask. Obtaining stability scores for all unlabeled images, we rank the entire unlabeled set according to these scores. 
Since IAN is a complete structure with one left and one right, we use its prior structure information to select the top K data with the highest stability score and the number of connected domains equal to 2 to generate pseudo labels. 
After that, we use the original labeled data and pseudo-labeled data to jointly supervise the training of the student model, and update the student model to a new iteration of the teacher model.
\subsubsection{Strong data augmentations.}

Weak or basic augmentations employed in conventional fully supervised semantic segmentation, including random rotation, resizing, brightness, cropping, and flipping. 
We inject SDA on pseudo-labeled images before training the student model to alleviate overfitting noisy labels and decouple similarity predictions between teacher and student, including color, noise, and painting jitter. 
In the prediction stage, test-time augmentation predictions are performed on all unlabeled images, including rotation, cropping, and flipping.

\section{Experiments}
\subsection{Dataset and Evaluation Metrics}
We conduct experiments on the 2023 ToothFairy dataset. 
The training set consists of primary dataset and secondary dataset. 
Among them, the primary dataset has 153 CBCT images with both dense and sparse annotations, and the secondary dataset has 290 CBCT images with only sparse annotations.
The test set contains 50 more cbct images and will not be publicly released even after the end of the challenge.

The evaluation measures consist of two accuracy measures: Dice Similarity Coefficient (DSC) and 95\% Hausdorff Distance (HD95), and two running efficiency measures: the maximum used memory (Mem), and the total execution time (Time) for all cases. 
The accuracy measures will be firstly used to compute the ranking. 
If ranks are equal, Mem and Time will be calculated to break ties.

\subsection{Implementation Details}
\subsubsection{Training set reconstruction}
After filtering out 2 cases with disconnected IAN and 40 cases with blurred ends, we reconstruct the training set as 111 labeled images and 332 unlabeled images.
\subsubsection{Selective re-training strategy}
Reliable images are measured by three checkpoints that are saved uniformly over 1/3, 2/3, 3/3 total iterations during training. 
We regard the top 100 images with the highest score and the number of connected domains as 2 with a meanDice score greater than 0.9 as reliable images, and the rest as unreliable images. 
We performed two pseudo-label update iterations, training the final model with 111 labeled data and 200 pseudo-labeled data.
\subsubsection{Training procedure and environments}
We use the same training strategy for both teacher and student model. 
Specifically, the batch size is set as 2 and the batch normalization (BN) is applied. 
Moreover, stochastic gradient descent (SGD) is used as the optimizer. 
The initial learning rate is set to be 0.01 and the total training epochs are 1000. 
The loss function is sum of cross entropy loss and dice loss. We implement our framework with PyTorch based on a single NVIDIA GeForce RTX 3090 GPU with 24 GB memory.

\subsection{Results}
The quantitative results are shown in Table \ref{metric}. 
It can be found that the proposed method can achieve very promising results on IAN segmentation as the number of iterations increases. 
In the final test phase with 50 cases of data, our method achieved the average dice similarity coefficient (DSC) of 0.7956, and 95\% hausdorff distance (HD95) of 4.4905 under 2 iterations, and won the challenge The first place, verified the effectiveness of our framework.

\begin{figure*}[]
	\centering
	\includegraphics[width=\columnwidth]{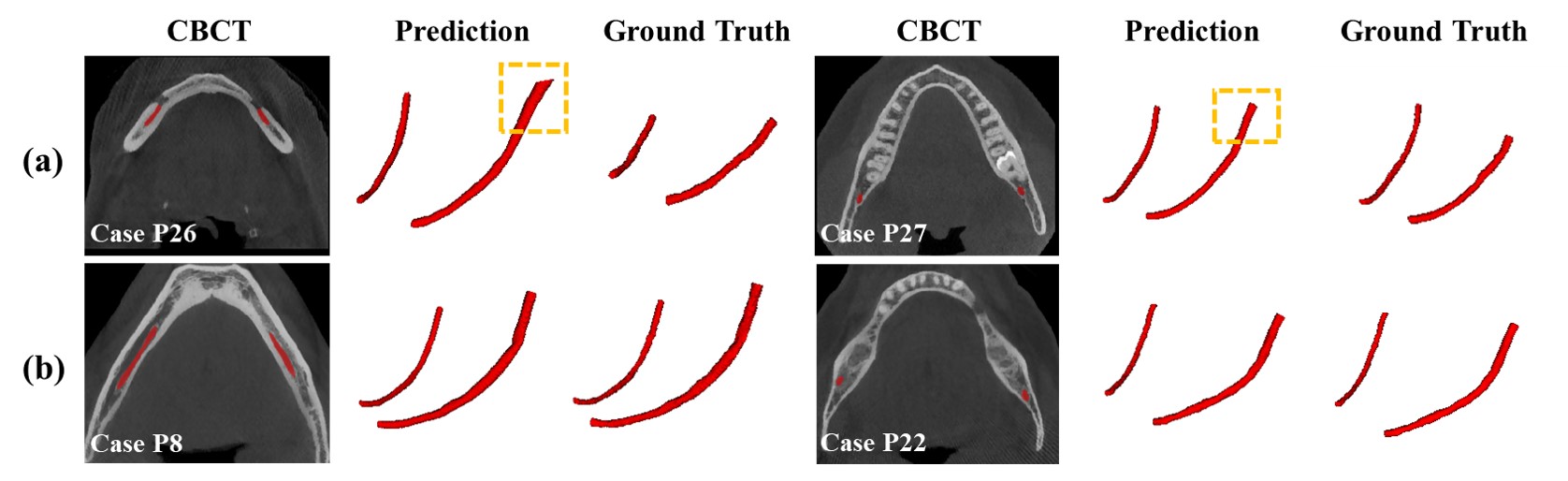}
	\caption{Qualitative results on hard (case P26 and P27) and easy (case P8 and P22) examples.}
	\label{visual}
\end{figure*}

Fig.\ref{visual} presents some hard and easy examples on the validation set. For Cases P26 and P27, our method segments more parts at both ends than the densely labeled results, which is caused by the blurred boundaries. 
For Case P8 and P22, our method obtains good segmentation results.

\begin{table}[!h]
    \centering
    \caption{The quantitative evaluation on preliminary and final test phase.}
    \label{metric}
    \setlength\tabcolsep{3.5mm}{
    \begin{tabular}{cccc}
    \hline
    \toprule[0.5pt]
    Iteration & Data Size                & DSC(\%)       & HD(mm)        \\ \hline
    \multicolumn{4}{c}{Preliminary Phase}                                \\ \hline
    0         & 111 labeled              & 0.7720 $\pm$ 0.0434 & 3.2472 $\pm$ 0.4282 \\
    1         & 111 labeled + 100 pseudo & 0.7832 $\pm$ 0.0436 & 2.6882 $\pm$ 0.4008 \\
    2         & 111 labeled + 200 pseudo & 0.7885 $\pm$ 0.0478 & 2.6310 $\pm$ 0.3420 \\ \hline
    \multicolumn{4}{c}{Final Test Phase}                                 \\ \hline
    2         & 111 labeled + 200 pseudo & 0.7956 $\pm$ 0.0931 & 4.4905 $\pm$ 6.0807 \\ \hline
    \toprule[0.5pt]    
    \end{tabular}}
    \end{table}

\section{Conclusion}
In this paper, we propose a self-training based semi-supervised framework for automatic and efficient IAN segmentation in CBCT images. 
We adopt a selective re-training strategy based on IAN connectivity to obtain reliable pseudo masks from densely labeled data. 
Furthermore, we use strong data augmentation (SDA) and test-time augmentation (TTA) on data that is not densely annotated to decouple similarity predictions between teacher and student models. 
The results show that our method has a good segmentation performance and won the first place in the Miccai ToothFairy2023 Challenge.


\bibliographystyle{ieeetr}
\bibliography{reference.bib}  
\end{document}